\documentclass[sigconf]{acmart}
\renewcommand\footnotetextcopyrightpermission[1]{} % removes footnote with conference information in first column
\usepackage{amsfonts,amsthm,bm}
\usepackage{float}              % For float arguments in Figures
\usepackage{graphicx}           % For Figures to be added
\usepackage{multirow}           % Multiple rows

\AtBeginDocument{%
  \providecommand\BibTeX{{%
    \normalfont B\kern-0.5em{\scshape i\kern-0.25em b}\kern-0.8em\TeX}}}

\acmSubmissionID{5}

\makeatletter
\def\@copyrightspace{\relax}
\makeatother
\begin{document}

\title[A Self-Supervised FMA Loss and CA Finetuning to Efficiently Improve the Robustness of CNNs]{A Self-Supervised Feature Map Augmentation (FMA) Loss and Combined Augmentations Finetuning to Efficiently Improve the Robustness of CNNs}

\author{Nikhil Kapoor}
%\email{nikhil.kapoor@volkswagen.de}
\affiliation{%
  \institution{Volkswagen AG}
  \streetaddress{Berliner Ring 2}
  \city{Wolfsburg}
  \country{Germany}
  \postcode{38440}
}

\author{Chun Yuan}
%\email{yuanchun0504@gmail.com}
\affiliation{%
  \institution{Volkswagen AG}
  \streetaddress{38440 Wolfsburg}
  \city{Wolfsburg}
  \country{Germany}}

\author{Jonas Löhdefink}
%\email{j.loehdefink@tu-bs.de}
\affiliation{%
  \institution{Technische Universität Braunschweig}
  \city{Braunschweig}
  \country{Germany}
}

\author{Roland Zimmermann}
%\email{roland.zimmermann@stud.uni-goettingen.de}
\affiliation{%
 \institution{Volkswagen AG}
 \streetaddress{Berliner Ring 2}
 \city{Wolfsburg}
 \country{Germany}}

\author{Serin Varghese}
%\email{john.serin.varghese@volkswagen.de}
\affiliation{%
  \institution{Volkswagen AG}
  \streetaddress{Berliner Ring 2}
  \city{Wolfsburg}
  \country{Germany}}

\author{Fabian Hüger}
%\email{fabian.hueger@volkswagen.de}
\affiliation{%
  \institution{Volkswagen AG}
  \streetaddress{Berliner Ring 2}
  \city{Wolfsburg}
  \country{Germany}}
 
\author{Nico Schmidt}
%\email{nico.maurice.schmidt@volkswagen.de}
\affiliation{%
  \institution{Volkswagen AG}
  \streetaddress{Berliner Ring 2}
  \city{Wolfsburg}
  \country{Germany}}
 
\author{Peter Schlicht}
%\email{peter.schlicht@volkswagen.de}
\affiliation{%
  \institution{Volkswagen AG}
  \streetaddress{Berliner Ring 2}
  \city{Wolfsburg}
  \country{Germany}}
 
\author{Tim Fingscheidt}
%\email{t.fingscheidt@tu-bs.de}
\affiliation{%
  \institution{Technische Universität Braunschweig}
  \city{Braunschweig}
  \country{Germany}
}

\renewcommand{\shortauthors}{Kapoor et al.}

\begin{abstract}
  Deep neural networks are often not robust to semantically-irrelevant changes in the input. 
   In this work we address the issue of robustness of state-of-the-art deep convolutional neural networks (CNNs) against commonly occurring distortions in the input such as photometric changes, or the addition of blur and noise. 
   These changes in the input are often accounted for during training in the form of data augmentation. We have two major contributions: First, we propose a new regularization loss called feature-map augmentation (FMA) loss which can be used during finetuning to make a model robust to several distortions in the input. Second, we propose a new combined augmentations (CA) finetuning strategy, that results in a single model that is robust to several augmentation types at the same time in a data-efficient manner. We use the CA strategy to improve an existing state-of-the-art method called stability training (ST). Using CA, on an image classification task with distorted images, we achieve an accuracy improvement of on average \textbf{8.94\%} with FMA and \textbf{8.86\%} with ST absolute on CIFAR-10 and \textbf{8.04\%} with FMA and \textbf{8.27\%} with ST absolute on ImageNet, compared to \textbf{1.98\%} and \textbf{2.12\%}, respectively, with the well known data augmentation method, while keeping the clean baseline performance.
\end{abstract}



\maketitle
\pagestyle{plain}
\section{Introduction}

Over the past few years deep neural networks (DNNs) have shown impressive performance on a variety of computer vision tasks such as image classification~\cite{Krizhevsky2012, He2015, Mahajan2018}, object detection~\cite{Girshick2015, Redmon2015, He2017}, semantic segmentation~\cite{Chen2017, Zhu2019, Wang2019, Loehdefink2019}, etc. However, recent works have demonstrated that these state-of-the-art networks are not robust to small changes in the input~\cite{Azulay2018, Hendrycks2019, Rodner2016, Engstrom2019, Bunne2018, Baer2019, Fawzi2015}. These small changes in the input, also called distortions, can be of various types, e.g., photometric changes (brightness, saturation, etc.)~\cite{Zhang2018} or noise (Gaussian, salt and pepper (SAP) noise, etc.)~\cite{Fawzi2017}.
% Figure 1: Examples of FMA output
\begin{figure}[t!]
    \centering
    %\resizebox{0.5\linewidth}{!}{
    \includegraphics[width=\linewidth]{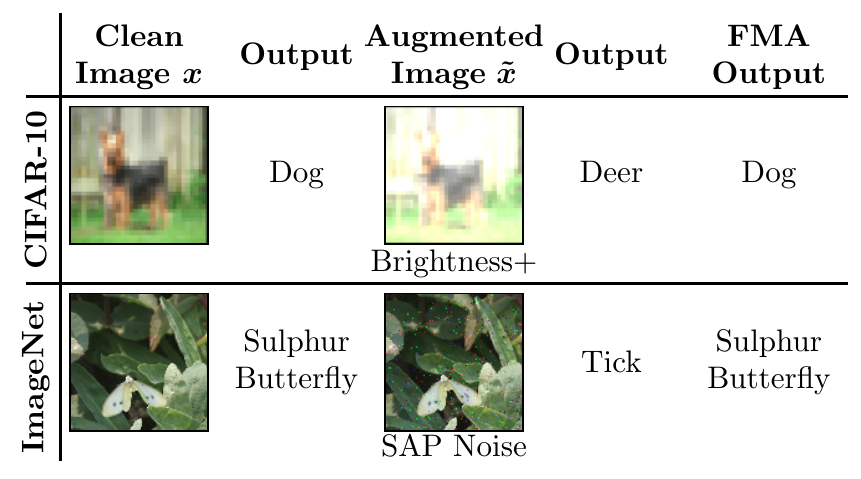}
    \caption{Examples of two commonly occurring distortions leading to misclassifications. Top row (from left): Example of clean image from CIFAR-10 correctly classified as \textit{dog}, gets misclassified as \textit{deer} when the \textit{brightness} is increased. Our proposed feature map augmentation (FMA) loss overcomes this misclassification providing \textit{dog} again. Bottom row (from left): Similar example from ImageNet of a \textit{butterfly} image falsely classified as a \textit{tick} when the augmentation type is salt and pepper (\textrm{SAP}) noise. The FMA loss overcomes this mistake.}
    \label{fig:FMA_improvement}
\end{figure}
In the real world, deviations from the training set distribution are to be expected. For instance, varying light conditions might affect the contrast and brightness of an image. Blurring of images might occur due a to shaky camera, bad weather conditions such as fog, rain, or simply incorrect camera settings (higher ISO, low shutter speed, etc.). Such changes in the input might lead to a wrong output of the model, as shown in Figure \ref{fig:FMA_improvement}. In order to safely deploy neural networks in safety-critical situations such as autonomous cars or in bio-medical applications, it is vital to ensure high accuracy on the original task as well as high robustness of the output to small changes of the input.

Since there can be many different types of deviations arising due to many different reasons, it is quite challenging to train truly robust models that are invariant to all possible input changes. Traditionally, data augmentation is performed in order to overcome this challenge~\cite{Wong2016}. This means that augmented images are added into the training set and the model is finetuned on this extended training set while keeping the original loss. However, there seems to be no clear understanding on how effective this approach truly is. It also suffers from the obvious downside of increased computational expense as the size of the dataset increases proportionally to the number of augmentations. Azulay \textit{et al.}~\cite{Azulay2018} also show that data augmentation merely leads to an increased invariance to augmentation types only for images that look very similar to typical training set images. This leads us to the question: \textit{"Can we improve on data augmentation and come up with a better way to ensure robustness of convolutional neural networks to commonly occurring image corruption types without the downside of increased computational expense?"} We aim to answer this question in the scope of this paper.

For this, we propose a novel feature map augmentation (FMA) loss as well as a new combined augmentation (CA) training strategy that aims at increasing robustness of convolutional neural networks to a pre-defined set of commonly occurring augmentation types in a data-efficient manner. Our method outperforms data augmentation by a large margin over a range of augmentation types on two different classification datasets, namely ImageNet~\cite{Krizhevsky2012} and CIFAR-10~\cite{Krizhevsky2009}, while maintaining its original task performance. We test our approach over five different augmentation types as shown in Table~\ref{tab:augmentation_parameters}. 
\noindent{In summary, our contributions are as follows:}
\begin{enumerate}
    \item We propose an \textit{additional feature map augmentation (FMA) loss term} that aims at making any given pre-trained convolutional neural network (CNN) model robust to a pre-determined set of input distortions using only a few \textit{subsequent} epochs of finetuning.
    
    \item For training a model with multiple augmentation types at the same time, we propose a new data-efficient \textit{combined augmentation (CA)} training strategy and use this to additionally improve on an existing state-of-the-art method for robust training.

    \item Finally, we demonstrate that when compared to data augmentation, our finetuned model is \textit{significantly more robust to multiple augmentation types at the same time} and also keeps its original classification accuracy.
\end{enumerate}

%Table 2: Augmentation parameters
\begin{table}[t!]
    \centering
    \caption{Augmentation parameters $\bm{\phi}_n$ that lead to a roughly $10\%$ absolute drop in validation performance for the \texttt{VGG-16} baseline model trained on CIFAR-10 and ImageNet datasets.}
    \begin{tabular}{c|ll}
        \multicolumn{2}{c}{\textbf{Augmentation}} & \textbf{Parameters $\bm{\phi}_n$} \\
        \toprule
        \multirow{7}{*}{\rotatebox[origin=c]{90}{\textbf{CIFAR-10}}}& Brightness$+$ &$\Delta = 0.39$ \\
        & Brightness$-$ &$\Delta = -0.36$ \\
        & Saturation$+$ &$\alpha = 6.0$ \\
        & Saturation$-$ &$\alpha = 0.0$ \\
        & Gaussian noise &$\mu = 0.0, \sigma = 0.075$ \\
        & Gaussian blurring & $s = 3.0, \mu = 0.0, \sigma = 0.675$ \\
        & Additive SAP noise & $p = 0.025, q = 0.5, \rho=0.5$ \\
        \bottomrule 
        \multirow{7}{*}{\rotatebox[origin=c]{90}{\textbf{ImageNet}}}& Brightness$+$ &$\Delta = 0.43$ \\
        & Brightness$-$ &$\Delta=-0.32$ \\
        & Saturation$+$ &$\alpha = 4.0$ \\
        & Saturation$-$ &$\alpha = 0.2$ \\
        & Gaussian noise &$\mu = 0.0, \sigma = 0.08$ \\
        & Gaussian blurring &$s = 3.0, \mu = 0.0, \sigma = 1.175$ \\
        & Additive SAP noise &$p = 0.01, q = 0.7, \rho = 0.7$ \\
        \bottomrule
    \end{tabular}
    \label{tab:augmentation_parameters}
\end{table}

\section{Related Work}
\label{sec:Related Work}

This section highlights existing work on data augmentation, other robustness enhancement techniques in general, and stability training in particular.

\subsection{Data augmentation}
Data augmentation is common practice in neural network training where training samples are augmented with different augmentation types and the network is trained on this extended data set~\cite{Ho2019, Wang2017}. Although it helps increase generalization, the computational complexity also increases. For the sake of this paper, we term this method as augmentation training (AT). Several approaches have been proposed to selecting clever augmentation policies such as Autoaugment~\cite{Cubuk2019}, AugMix~\cite{Hendrycks2020}, Randaugment~\cite{Cubuk2020} etc., however most of these approaches tend to add additional computational overhead of searching for an effective augmentation strategy. More so, while these approaches tend to increase generalization, they are not particularly efficient in improving robustness to a held out set of augmentations. In contrast, we aim at improving robustness of an already well trained model, given a set of pre-defined relevant augmentations.

\subsection{Robustness enhancements}
In order to improve classifier stability, Vasiljevi \textit{et al.} ~\cite{Vasiljevic2016} finetuned on blurred images. To generalize to other blurs, they found that it is not enough to finetune on one type of blur to generalize to other blur types. On the other hand, Rosza \textit{et al.}~\cite{Rozsa2018} proposed a simple training technique called batch-adjusted network gradients (BANG) that does not need any additional training data to improve robustness. They propose a slight variation of batch normalization by balancing weight updates which inherently increases robustness in general by smoothing the decision boundaries. However, their approach is not self-adaptive to other models and tasks. Geihros \textit{et al.}~\cite{Geirhos2019} proposed Stylized ImageNet, where clean images were converted to different styles/textures, such as canvas paints and the model was trained on these stylized images in addition to clean images. In order to evaluate corruption robustness, Hendrycks \textit{et al.}~\cite{Hendrycks2019} proposed a public benchmark of 15 corruption types at 5 different severities on CIFAR-10 and ImageNet dataset, however, this dataset is only meant to be tested for methods that do not explicitly train on the same augmentations. For this reason, we do not use their benchmark for evaluation of our results.

\subsection{Stability training (ST)}
Zheng \textit{et al.}~\cite{zheng2016} introduced an additional regularization loss which penalizes the prediction difference of the softmax output of clean and perturbed images. Additionally, they propsoed to train only on images augmented with Gaussian Noise (GN) as a means of improving robustness in a general manner to many augmentation types. However, on re-implementation of their method, we could not confirm this to be true: We noticed that training with GN only helped improve robustness to GN and other noise types such as salt and pepper, with a corresponding robustness decrease in other augmentation types. Hence, to improve further on ST~\cite{zheng2016} and conventional data augmentation (AT)~\cite{Wong2016}, we propose a new combined augmentation (CA) training strategy which will be discussed later. 

\section{Feature Map Augmentation (FMA) Loss \& Training Strategy}
\label{sec:FMA}

% Figure 2: FMA Method Overview
\begin{figure}[t!]
    \centering
    %\resizebox{0.5\linewidth}{!}{
    \includegraphics[width=\linewidth]{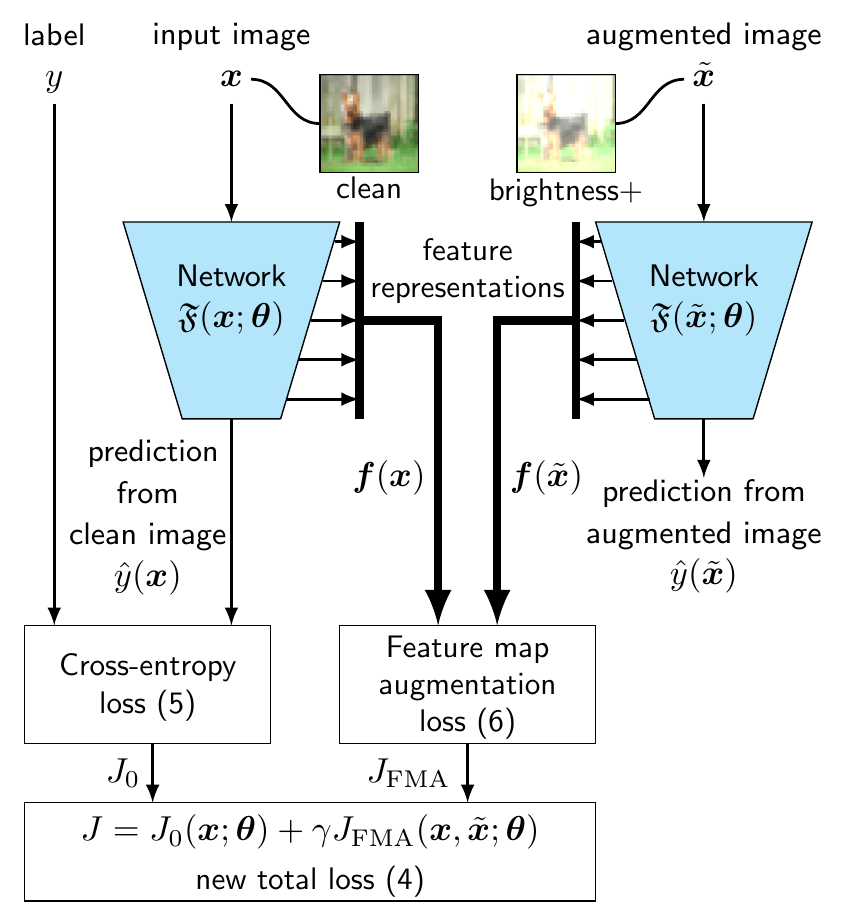}%}
    \caption{\textbf{Finetuning with new total loss}, including \textbf{feature map augmentation (FMA) loss}. A CNN model pre-trained using the original task loss $J_0$~(\ref{eq:L_0}) can be finetuned using the loss $J$~(\ref{eq:loss_function}) including the new regularization loss called feature map augmentation (FMA) loss $J_\mathrm{FMA}$~(\ref{eq:L_FMA}). The FMA loss is computed by a self-supervised regularization of differences of feature activation maps of all layers to clean and augmented image pairs. The hyper-parameter $\gamma$ controls the trade-off between clean and augmented accuracy.}
    \label{fig:FMA_overview}
\end{figure}
In this section, we first describe the intuition behind the idea of the feature map augmentation (FMA) loss. We then present details of our method and show how this can be used to stabilize feature embeddings.
% Figure 3: Augmentation Visualization
\begin{figure*}[t]
	\centering
	\includegraphics[width=\linewidth]{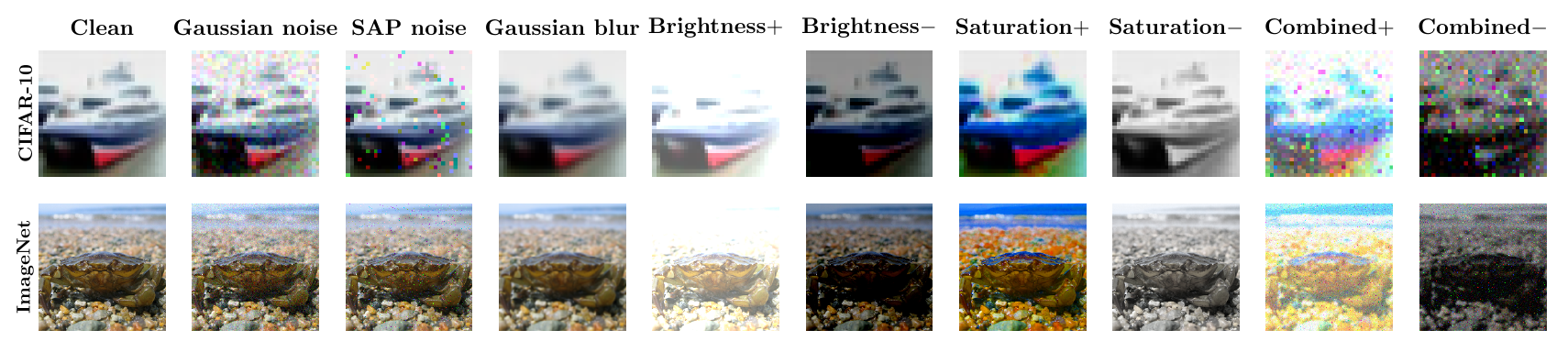}
	\caption{Qualitative \textbf{visualization of augmentation types} as introduced in Table~\ref{tab:augmentation_parameters}. '$+$' and '$-$' here denote increase and decrease, respectively. Combined$+$ means all augmentations applied at the same time except Brightness$-$ and Saturation$-$. Combined$-$ on the other hand means all augmentations applied at the same time, except Brightness$+$ and Saturation$+$. Each augmentation leads to a drop of $10\%$ absolute performance on the validation set, except Saturation$+/-$ and Combined$+/-$.
	\textit{Top row:} CIFAR-10 clean and augmented example image. \textit{Bottom row:} ImageNet clean and augmented example image.}
	\label{fig:augmentation_visualization}
\end{figure*}
\subsection{Intuition}
In an ideal world, a CNN should respond similarly to a clean and an augmented image as long as the semantic content of the images remains the same, i.e., the network is expected to be augmentation-invariant. Therefore, we would hope that the feature activation map of individual layers should also remain the same, given a clean and its corresponding augmented input. If not, we can assume that the filters corresponding to the deviations in feature maps are sensitive to augmentations. In consequence, we will propose a new feature map augmentation (FMA) loss which regularizes the normalized difference in feature maps between a clean and an augmented image. We expect that this would increase the model robustness as the robustness objective is now made explicit and the model is optimized towards achieving this goal.

\subsection{Robustness Objective}
In this section we define our robustness objective that we optimize for. First, however, we define some mathematical~notation. 

Let $\mathcal{A}=\{A_1, \dots, A_n, \dots, A_N\}$ be a set of $N$ augmentation types, as also shown in Table~\ref{tab:augmentation_parameters}. We define $\bm{x}~\in~\mathbb{G}^{H \times W \times C}$ as an image of dataset \textbf{$\mathcal{X}$} with $\mathbb{G}~=~[0,1]$ being the set of gray values, image height $H$, image width $W$, and number of color channels $C$. The image $\bm{x}$ is fed into a neural network $\mathfrak{F}(\bm{x};\bm{\theta})$ having the network parameters $\bm{\theta}$. The neural network $\mathfrak{F}(\bm{x};\bm{\theta})$ consists of several layers $\ell \in \mathcal{L}=\{1, 2, \dots,L\}$, each having an output feature map $\bm{f}_\ell(\cdot) \in \mathbb{R}^{H_{\ell} \times W_{\ell} \times C_{\ell}}_{\geq 0}$ (assuming a ReLU activation function) with the height $H_{\ell}$, width $W_{\ell}$ and number of feature maps $C_{\ell}$, and $\mathbb{R}_{\geq 0} =\{x \in \mathbb{R} \mid x \geq 0\}$.

Given this notation, we can write the overall computation of the neural network $\mathfrak{F}(\bm{x};\bm{\theta})$ as follows:
\begin{equation}
    \hat{y} = \mathfrak{F}(\bm{x};\bm{\theta}) = o(\bm{f}_L(\bm{f}_{L-1}(\dots (\bm{f}_{2}(\bm{f}_{1}(\bm{x}))))))
\end{equation}
where $\bm{f}_{\ell}$ denotes the feature map tensor of layer $\ell$, where $\ell~\in~\mathcal{L}$, and $o(\cdot)$ denoting the output layer providing a scalar prediction. For each augmentation type $A_n$, we can compute the corresponding augmented image as follows:
\begin{equation}
   \Tilde{\bm{x}} = \bm{\delta}_n(\bm{x}) \in \mathbb{G}^{H \times W \times C},
    \label{eq:x_dash}
\end{equation}
where $\bm{\delta}_n(\cdot)$ is the augmentation function with specific parameters (see Table \ref{tab:augmentation_parameters}) $\bm{\phi}_{n} \in \bm{\mathcal{\phi}} = \{\bm{\phi}_{1},\dots,\bm{\phi}_{N}\}$, corresponding to the augmentation type $A_{n}~\in~\mathcal{A}$. Irrespectively of the applied augmentation type, function $\bm{\delta}_n(\cdot)$ always performs clipping of each pixel to enforce $\bm{\delta}_n(\cdot)~\in~\mathbb{G}^{H \times W \times C}$. This allows handling zero-mean noise as well. In order to make our model robust to all the augmentations, we want to ensure that the feature maps are similar for both clean as well as distorted images for all the augmentations. The robustness objective is therefore,
\begin{equation}
    \forall (\bm{x},\Tilde{\bm{x}}) : \bm{f}_{\ell}(\bm{x}) \approx \bm{f}_{\ell}(\Tilde{\bm{x}}), \ell \in \mathcal{L}
    \label{eq:robustness_objective}
\end{equation}
with $\Tilde{\bm{x}}$ as in~(\ref{eq:x_dash}). Given an existing training objective $J_0$ on the original task (e.g., classification), an input image $\bm{x}$ and a perturbed copy $\Tilde{\bm{x}}$, we can implement the new total loss $J$ for finetuning with the robustness objective~(\ref{eq:robustness_objective}) as:
\begin{equation}
    J(\bm{x}, \Tilde{\bm{x}}; \bm{\theta}) = J_0(\bm{x}; \bm{\theta}) + \gamma J_{\mathrm{FMA}}(\bm{x}, \Tilde{\bm{x}}; \bm{\theta}),
    \label{eq:loss_function}
\end{equation}
where $\gamma$ controls the strength of the regularization term $J_{\mathrm{FMA}}$. In terms of classification, the $J_0$ term can be a standard cross-entropy loss
\begin{equation}
    J_{0}(\bm{x}; \bm{\theta}) = -\sum_{j \in \mathcal{J}}{\hat{y}}_{j} \log P(y_{j}|\bm{x}; \bm{\theta}),
    \label{eq:L_0}
\end{equation}
where the index $j \in \mathcal{J}$ runs over the number of classes and ${\hat{y}}_{j} \in \{0,1\}$ is a binary indicator being 1 if the predicted class label $j$ is the correct classification and 0 otherwise. The new loss term in (\ref{eq:loss_function}) can then be defined as
\begin{equation}
        J_{\mathrm{FMA}}(\bm{x}, \Tilde{\bm{x}};\bm{\theta}) = \frac{1}{|L|}\sum_{\ell \in \mathcal{L}}\frac{1}{\kappa_{\ell}}\Bigg|\Bigg|\frac{\bm{f}_{\ell}(\bm{x}) - {\bm{f}}_{\ell}(\Tilde{\bm{x}})}{{\overline{\bm{f}_{\ell}(\bm{x})}}}\Bigg|\Bigg|^{2},
    \label{eq:L_FMA}
\end{equation}
where $||\cdot||^2$ denotes the squared $L_2$ norm, and $\overline{\bm{f}_{\ell}(\bm{x})}$ is the mean of all entries in the respective feature map tensor. Here, $\kappa_{\ell}=~H_\ell~\times~W_\ell~\times~C_\ell$ corresponds to the dimension of the feature map in layer $\ell$. 

\subsection{Finetuning with FMA}
\label{subsec:FMA_training}
We now demonstrate our two-stage training approach using the new FMA loss~(\ref{eq:L_FMA}) as part of the total loss~(\ref{eq:loss_function}) in finetuning. An overview  is also shown in Figure~\ref{fig:FMA_overview}.

\subsubsection{Baseline model training}
We train a baseline model on the original task using clean images and their labels. This could be, for example, on a cross-entropy loss~(\ref{eq:L_0}) for a classification task.

\subsubsection{Compute augmentation parameters}
Next, we need a mechanism to compute the strength of the augmentations that we want to make our model robust against. For the sake of simplicity, and fair comparison at the end, we choose these parameters such that the performance of the model on the augmented validation set drops by roughly $10\%$ absolute. Once this is attained, the parameters $\bm{\phi}_{n}$ for $\bm{\delta}_n(\cdot)$ are frozen. We list these parameters in Table~\ref{tab:augmentation_parameters} and  visualize their qualitative influence in Figure~\ref{fig:augmentation_visualization}.

\subsubsection{Robustness finetuning with FMA loss}
Now, we augment the clean images with several distortions {\textit{at once}} (performing final clipping)
\begin{equation}
\label{eq:augmentation_addition}
\Tilde{\bm{x}} = \bm{\delta}_n(\bm{\delta}_{n-1} \dots (\bm{\delta}_{1}(\bm{x})),    
\end{equation}
where the index $n \in \mathcal{N}$ runs over the number of augmentation types. Lastly, we finetune the baseline model of stage 1 on the new total loss function as defined in~(\ref{eq:loss_function}). Here, the labels for the augmented images are not needed as only the feature map activations are compared for computing the new FMA loss term as defined in~(\ref{eq:L_FMA}). The hyperparameter $\gamma$ is computed using grid search as in~\cite{Bergstra2012}.

\section{Implementation Details}
\label{sec:Network, Augmentations, and Training Protocols}
% Table 2: Experiment IA Results
\begin{table*}[t!]
  \centering
  \caption{\textbf{Accuracy values for individual augmentations both in training (IA strategy) and test} (Section~\ref{subsec:Robustnesss_individual_augmentations}) on the CIFAR-10 validation set. For each augmentation type, a baseline model is finetuned with three different methods, i.e., augmentation training (AT), stability training (ST), or by our new FMA-based total loss~(\ref{eq:L_FMA}). The evaluation is performed on both clean and augmented validation sets for the same augmentation type that it was finetuned for. $\mathrm{ACC}_1$ and $\mathrm{ACC}_2$ refer to accuracy of a model evaluated on the clean validation set before and after robust training, respectively. Similarly, $\widetilde{\mathrm{ACC}_1}$ and $\widetilde{\mathrm{ACC}_2}$ refer to the accuracy of a model evaluated on the augmented validation set before and after robust training, respectively.}
%\resizebox{0.9\linewidth}{!}{
\begin{tabular}{l|c|c|ccccccc|c}
    \toprule 
          \multicolumn{2}{l|}{\textbf{Accuracy}} & \textbf{Training} &        \multicolumn{7}{c|}{\textbf{Augmentation Type $A_n$}} & \textbf{Average} \\
         \cline{4-10}
          \multicolumn{2}{l|}{} & \textbf{Method} &
          \textbf{B+} & \textbf{B-} & \textbf{GB} & \textbf{GN} & \textbf{SAP} & \textbf{S$+$} & \textbf{S$-$} & \textbf{Improvement} \\
          \hline 
         \multirow{4}{*}{\rotatebox{90}{\textbf{\footnotesize{Clean}}}} & $\mathrm{ACC}_1$ & & 89.82\% & 89.82\% & 89.82\% & \textbf{89.82\%} & 89.82\% & 89.82\% & 89.82\% & $\mathrm{ACC}_2 - \mathrm{ACC}_1$ \\
         \cline{2-11} 
           & \multirow{3}{*}{$\mathrm{ACC}_2$} & AT & 90.35\% & 90.26\% & 90.00\% & 89.32\% & 89.51\% & 90.06\% & 90.17\% & 0.13\% \\
          & & ST  & \textbf{90.58\%}  & 90.32\% & \textbf{90.56\%} & 89.79\%  & \textbf{89.94\%} & \textbf{91.40\%} & \textbf{90.76\%} & \textbf{0.65\%} \\
          & & FMA & 90.20\%  & \textbf{90.33\%} & 90.23\% & 88.83\% & 89.00\% & 90.47\% & 89.43\% & $-$0.03\% \\
          \midrule  
          \multirow{4}{*}{\rotatebox{90}{\textbf{\footnotesize{Augmented}}}} & $\widetilde{\mathrm{ACC}_1}$ & & 80.13\% & 80.13\% & 81.06\% & 81.17\% & 80.73\% & 80.47\% & 84.03\% & $\widetilde{\mathrm{ACC}_2} - \widetilde{\mathrm{ACC}_1}$ \\
          \cline{2-11} 
          & \multirow{3}[0]{*}{$\widetilde{\mathrm{ACC}_2}$} & AT & 87.80\% & 86.04\% & 89.17\% & 87.70\% & 86.79\% & 87.90\% & 87.42\% & 6.44\% \\
          & & ST & 88.50\%  & 86.97\% & \textbf{89.39\%} & \textbf{88.29\%} & \textbf{87.80\%} & 88.67\% & 87.93\% & 7.11\% \\
          & &  FMA & \textbf{88.69\%} & \textbf{87.47\%} & 89.32\% & 88.16\% & 87.54\% & \textbf{89.37\%} & \textbf{88.47\%} & \textbf{7.32\%} \\
          
       \bottomrule  
    \end{tabular}%}
  \label{tab:individual_experiments_results_cifar10}
\end{table*}
In this section, we first present implementation relevant details such as augmentation types, the network and datasets used with corresponding hyperparameters and then introduce our novel combined augmentation (CA) training strategy.

\subsection{Network and Dataset}
\label{subsec:Networks_and_Dataset}
Our experiments are performed for the image classification task on two well-known datasets, namely CIFAR-10~\cite{Krizhevsky2009} and a subset of ImageNet~\cite{Krizhevsky2012}. The CIFAR-10 dataset consists of $50,000$ training images and $10,000$ test samples being $32\times32$ color images sorted in $10$ classes. The ImageNet dataset (ILSVRC 2012) on the other hand consists of a total of $1.2$ million training images and $50,000$ validation and $150,000$ test samples being $224\times224$ color images sorted in $1000$ classes. For the sake of computational ease, we consider a subset of $200$ randomly chosen classes from the ImageNet dataset instead. This reduces the number of images to $240,000$ training images and $10,000$ test images, thereby accelerating our experiments significantly. For both of these datasets, we consider a standard \texttt{VGG-16}~\cite{simonyan2014} model pre-trained on ImageNet weights (downloaded from the official \texttt{tf-slim} repository\footnote{\url{https://github.com/tensorflow/models/tree/master/research/slim\#Pretrained}}). The model is adapted to both the datasets in terms of its input and output dimensions, and is trained using stochastic gradient descent with momentum optimizer for an additional $40$ epochs with varying learning rates $\eta_0=(10^{-2}, 10^{-4}, 10^{-6})$ for $(20, 10, 10)$ epochs, respectively. With this configuration, we achieve a baseline accuracy of \textbf{89.82\%} and \textbf{$79.77\%$} on the validation set of the CIFAR-10 and the ImageNet dataset, respectively. We use these baseline models for all our experiments.

\subsection{Augmentation Types and Parameters}
In this section, we explain the augmentation types (see Tables~\ref{tab:augmentation_parameters}) considered in this work and a few implementation details. Remember that pixel-wise clipping is finally performed in each of the augmentation functions $\bm{\delta}_n(\cdot)$ in (\ref{eq:x_dash}). We also introduce combinations of these augmentations that are used for later experiments. Combinations of augmentations always follow (\ref{eq:augmentation_addition}) with additional final pixel-wise clipping.
 
 \subsubsection{Photometric augmentation types} Distortions of this class changes occur mainly due to variations in lighting conditions. We consider two such changes, namely brightness, and saturation.
 \textit{Brightness (B)} can be changed by adding or subtracting a constant $\Delta$ to each of the RGB channels. \textit{Saturation (S)} changes can be implemented by multiplying the image's saturation in the hue, saturation and luminance (HSL) representation by a factor $\alpha$. 
 
 \subsubsection{Noise and blurring augmentation types}
 \textit{Gaussian noise (GN)} can occur due to sensor noise by poor illumination and/or high temperature, etc. \textit{Additive salt-and-pepper noise (SAP)} can occur due to bit errors in image transmission~\cite{Hendrycks2019}. \textit{Gaussian blurring (GB)} can occur due to out-of-focus images or incorrect camera configuration. For implementation, the Gaussian noise is zero-mean with standard deviation $\sigma$. In contrast to this, additive salt-and-pepper noise adds a binary noise with strength $\pm\rho$ to the image, where the probability of the noise is given by $p$ and the salt-to-pepper ratio by $q$. For the Gaussian blur, a zero-mean Gaussian kernel of size $s \times s$ with standard deviation $\sigma$ is convoluted over the image $\bm{x}$.
%Table 3: Combined training final model tested individually, compared with AT.
\begin{table*}[t!]
%FMA:
%cifar10-checkpoint-39
%imagenet:checkpoint-25
%adv:
%cifar10-checkpoint-28
    \centering
    \caption{\textbf{Accuracy values for models finetuned with combined augmentations (AT/CA, ST/CA and FMA/CA)} by using multiple augmentations simultaneously. The baseline model, as well as the three models after finetuning, are then evaluated with individual augmentations separately, as well as with clean images on the validation set. The strength of augmentations used in the combined training approach is such that most augmentation types individually lead to a performance drop of roughly 10\%. Therefore, when applied together, the combined accuracy drop is much more severe than 10\%.}
    %\resizebox{0.9\linewidth}{!}{
    \begin{tabular}{c|cccc|cccc}
        \toprule
        & \multicolumn{4}{c|}{CIFAR-10 validation set} &
        \multicolumn{4}{c}{ImageNet validation set} \\
        &\textbf{Baseline} & \textbf{AT/CA} & \textbf{ST/CA} &
        \textbf{FMA/CA} &\textbf{Baseline} & \textbf{AT/CA} & \textbf{ST/CA} &
        \textbf{FMA/CA}  \\
        \hline
        \textbf{Clean} & 89.82\% & 88.14\% & \textbf{89.91\%} & 88.97\%  & 79.77\% & 78.14\% & 81.01\% & \textbf{81.19\%}  \\
        \textbf{Brightness$+$ (B$+$)} & 80.13\% & 75.92\% & \textbf{82.32\%} & 82.23\% &  69.72\% & 64.89\% & \textbf{73.66\%} & 70.00\% \\
        \textbf{Brightness$-$ (B$-$)} & 80.13\% & 74.40\% & 81.14\% & \textbf{81.84\%}&   68.91\% & 65.12\% & 70.18\% & \textbf{70.77\%} \\
        \textbf{Gaussian blur (GB)} & 81.06\% & 80.31\% & 83.71\% & \textbf{85.02\%} &   70.59\% & 68.24\% & 70.72\% & \textbf{71.75\%} \\
        \textbf{Gaussian noise (GN)} & 81.17\% & \textbf{87.34\%} & 86.53\% & 86.94\% &  70.81\% & 75.34\% & 74.03\% & \textbf{75.44}\% \\
        \textbf{Additive SAP noise (SAP)} & 80.73\% & \textbf{86.94\%} & 85.72\% & 86.58\% & 72.32\% & \textbf{78.02\%} & 75.90\% & 76.99\% \\
        \textbf{Saturation$+$ (S$+$)} & 80.47\% & 73.71\% & \textbf{84.38\%} & 83.74\% &  67.30\% & 64.25\% & \textbf{68.63\%} & 66.98\% \\
        \textbf{Saturation$-$ (S$-$)} & 84.03\% & 82.03\% & \textbf{84.44\%} & 83.70\% &  64.15\% & 61.53\% & 66.87\% & \textbf{69.61\%} \\ 
        \textbf{Combined$+$} & 44.22\% & 54.62\% & \textbf{74.08\%} & 73.50\% &  18.46\% & 33.83\% & \textbf{58.06\%} & 54.93\% \\
        \textbf{Combined$-$} & 29.87\% & 48.10\% & 67.98\% & \textbf{68.59\%} &  13.47\% & 27.43\% & \textbf{39.15\%} & 38.27\% \\
        \hline
        \textbf{Average improvement} & --- & 1.98\% & 8.86\% & \textbf{8.94\%} &  --- & 2.12\% & \textbf{8.27\%} & 8.04\% \\
        \bottomrule
    \end{tabular}%}
    \label{tab:multi_augmentatations_individual_test}
\end{table*}
\subsection{Training Strategies}
\label{subsec:training_protocols}
We employ two training strategies for our experiments that are discussed next.

\subsubsection{Individual augmentation (IA) training strategy}
Individual augmentation refers to the strategy of augmenting each image with only a single augmentation type \textit{one at a time}. Hence, multiple augmentation types would mean multiple copies of the original image, each augmented with a different augmentation type. This increases the size of the dataset proportionally to the number of augmentation types considered. The strength of the augmentations is chosen such that the validation accuracy drops by roughly $10\%$. 

\subsubsection{Combined augmentation (CA) training strategy}
As the number of augmentation types increases, so does the dataset size. Training on this extended data set can involve high computational cost. In order to be data-efficient, we propose an alternate approach called CA training strategy. In contrast to IA, this means augmenting the same image with multiple augmentation types \textit{all at once}. Hence, as the number of augmentation types increases, the dataset size does not explode, and hence being more efficient. However, a naive combination of all augmentations at once can be counter-effective as some augmentation types (such as increase/decrease of brightness) can cancel each other out. We term such augmentations as \textit{mutually inverse augmentations}. In order to counter the inverse effect, we propose to create two sets of such augmentations following (\ref{eq:augmentation_addition}). We describe this in more detail now. However, before doing so, we define a few notations.

Consider the set of all individual augmentations denoted by $\mathcal{A} = \{\mathrm{B+}, \mathrm{B-}, \mathrm{S+}, \mathrm{S-}, \mathrm{GN}, \mathrm{SAP}, \mathrm{GB}\}$ where '$+$' and '$-$' indicate an \textit{increase} and \textit{decrease}, respectively. Assuming the set of \textit{non-inverse} augmentations defined as $\mathcal{A}' = \{\mathrm{GN}, \mathrm{SAP}, \mathrm{GB}\}$, we can reasonably group the \textit{inverse} augmentations as "Combined$+$", containing $\{\mathrm{B+}, \mathrm{S+}\} \cup \mathcal{A}'$, and "Combined$-$", containing $\{\mathrm{B-}, \mathrm{S-}\}~\cup~\mathcal{A}'$. Given such a grouping, during training, we simply alternate between Combined$+$ and Combined$-$ \textit{every epoch}, such that both the inverse augmentations do not cancel each other out in this way. One could also alternate images \textit{every batch}, however from our experiments, this led to sub-optimal results. Lastly, if we increase the number of epochs between the altering augmentation sets, this would lead to catastrophic forgetting~\cite{Kemker2017}. 

\section{Experiments and Results}
\label{sec:Experiments_and_Results}

The experiments in this section are split into two parts. The first part deals with testing our method against individual augmentations $\Tilde{\bm{x}}$ as defined in~(\ref{eq:x_dash}) and comparing our method with existing state-of-the-art robustness methods. The model is trained on an augmentation type, say $A_{1}$ and then tested on the same augmentation type during evaluation. The second part deals with evaluating the effectiveness of our combined augmentation (CA) training strategy on different methods. The aim is to investigate if training using CA helps improve robustness on individual augmentations separately. 

\subsection{Training With Individual Augmentations}
\label{subsec:Robustnesss_individual_augmentations}
As a first experiment, we investigate the increase in robustness of our model finetuned to seven individual augmentations (as shown in Table~\ref{tab:augmentation_parameters}) separately. We test our method with existing state-of-the-art robustness methods, namely augmentation training (AT) and stability training (ST) and show results in Table~\ref{tab:individual_experiments_results_cifar10}. For the sake of computational ease, we only test this on the CIFAR-10 dataset. For each method, 30 epochs of finetuning is performed. The baseline model has a validation accuracy of \textbf{89.82\%}. On top of this baseline model, we run in 21 additional finetunings (considering seven augmentation types, with three different training methods each). For each augmentation type, the strength of the augmentation is chosen such that the performance drop of 10\% on the validation set is attained, except in the case of Saturation$-$. This is because the saturation can be reduced only up to minimum $\alpha=0$ and this leads to a validation accuracy of \textbf{84.03\%} instead of 80\%. 

\subsubsection{Effect on clean data}
From the results shown in Table~\ref{tab:individual_experiments_results_cifar10}, all the three methods are either very close to the baseline performance or marginally better ($\mathrm{ACC_2}-\mathrm{ACC_1} \gtrsim 0$). We attribute this to the original loss $J_0$~(\ref{eq:L_0}) which is retained in all the three methods. However, surprisingly, we observe that AT does not lead to the best results on clean data, despite using the loss $J_0$~(\ref{eq:L_0}) at all times.

\subsubsection{Effect on augmented data}
Let $\widetilde{\mathrm{ACC}_1}$, $\widetilde{\mathrm{ACC}_2}$ denote the accuracy of the model evaluated on the augmented validation set before and after robust training, respectively. We analyze next the performance improvement on augmented data ($\widetilde{\mathrm{ACC}_2}-\widetilde{\mathrm{ACC}_1}$). We observe that with the new FMA loss, we obtain the best results with an average improvement of \textbf{7.32\%} absolute over all augmentation types, in comparison to  7.11\% and 6.44\% for ST and AT, respectively. These results indicate that our model is actually more robust to the augmentations that it was trained for. Interestingly, we managed to recover about 7\% of the augmented accuracy back with our method, keeping in mind that we started with about 10\% drop in augmented validation performance, while keeping the clean accuracy.
%Figure 4: Combined training tested on individual augmentations.
\begin{figure*}[t!]
	\centering
	\resizebox{0.75\linewidth}{!}{
	\includegraphics[width=\linewidth]{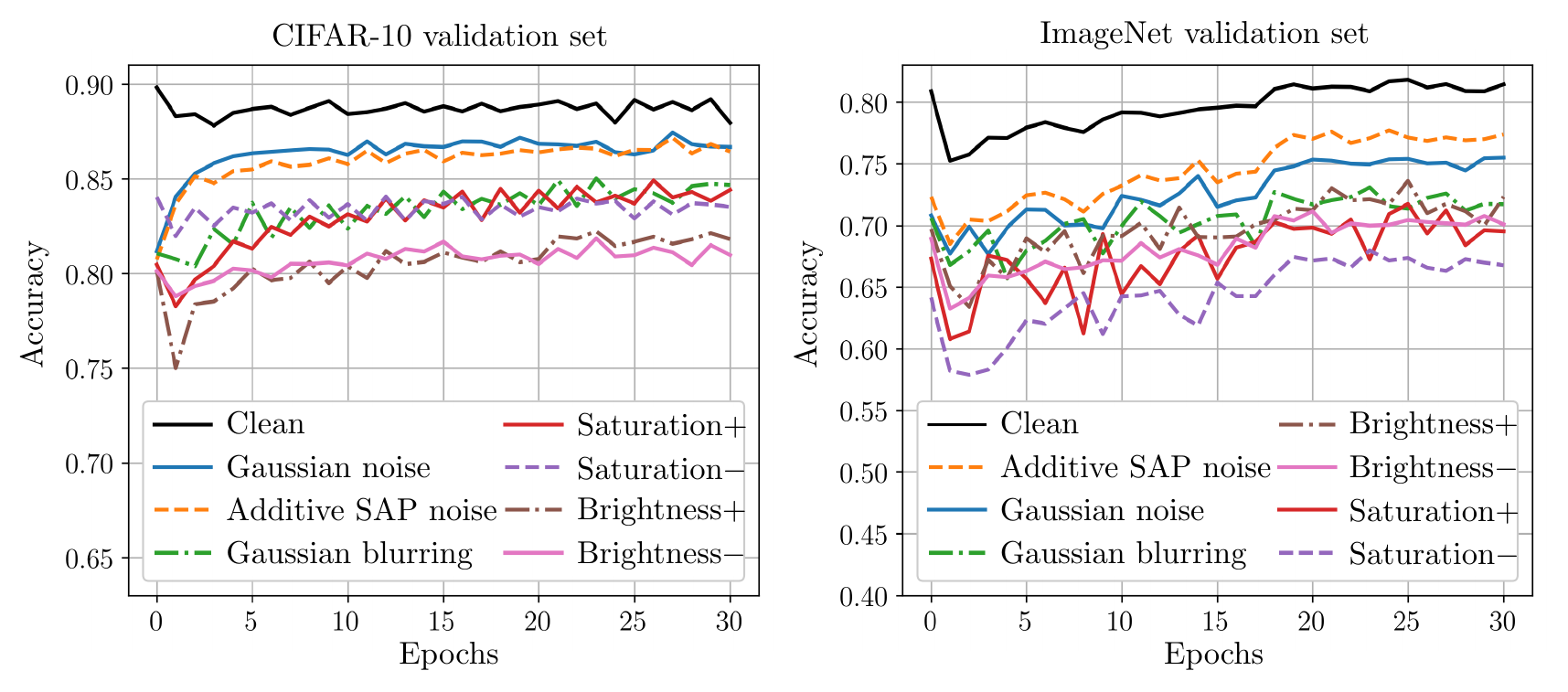}}
	\caption{\textbf{Accuracy during finetuning of our FMA/CA model} evaluated on individual augmentations (Table \ref{tab:augmentation_parameters}) independently. Results are shown for two datasets, CIFAR-10 and ImageNet.}
	\label{fig:combined_test_individual_CIFAR}
\end{figure*}
\subsection{Training With Multiple Augmentations}
\label{subsec:robustness_to_multiple_augmentations}

The second set of experiments investigates the combined augmentation (CA) training strategy (Section~\ref{subsec:training_protocols}). We find this very interesting, as we aim at having one model at the end of the training, which is robust to multiple augmentations at the same time, while keeping its original task performance. We test this on both CIFAR-10 and ImageNet validation sets. We investigate all three methods (AT, ST, and FMA) trained with CA, dubbed as AT/CA, ST/CA and FMA/CA, respectively. Although we could also test AT trained with IA training strategy as well, but we skip this as training our model with an 8x times dataset is computationally very expensive. We estimate a training time of 50 days for fine-tuning the baseline model for 30 epochs on the ImageNet subset dataset with these 7+1=8 augmentation types on a single \texttt{Nvidia GeForce GTX 1080Ti} GPU. This time is reduced to 6 days for AT/CA. Hence, in total, we perform six finetunings (three methods: FMA/CA, ST/CA and AT/CA for models trained on two datasets: CIFAR-10, and ImageNet). We first evaluate the improvement in augmented accuracies for all 6 final models quantitatively in Table~\ref{tab:multi_augmentatations_individual_test}. Then, we also visualize the same improvement in augmented accuracies at each step of training for FMA/CA in Figure~\ref{fig:combined_test_individual_CIFAR}.

\subsubsection{Effect on clean data and combined augmentations}
From Table~\ref{tab:multi_augmentatations_individual_test}, we first observe that the clean performance is more or less recovered with all the three methods. Surprisingly however, as also noticed from the previous experiment, AT fails to attain the best results on clean in comparison to ST and FMA on both datasets. We then compare models trained on the CIFAR-10 dataset and observe that the FMA and ST models when evaluated on Combined$+$, Combined$-$ have an impressive average absolute improvement of about \textbf{34\%}. On the other hand, the model trained using AT achieves an average improvement of about 14\%. Similarly, on the ImageNet dataset, we achieve an average improvement of about \textbf{31\%} for FMA and ST compared to about 15\% for AT.

\subsubsection{Effect on individual augmentations}
Next, we investigate the robustness improved on each augmentation type separately, even though during training, combined augmentations were used on individual images with CA. Absolute accuracies are reported in Table~\ref{tab:multi_augmentatations_individual_test} for all three methods AT, ST and FMA trained using CA. We notice that both FMA and ST methods help increasing augmented accuracies for majority of the augmentations. On the other hand, on both datasets, AT performs worse than the baseline model on all augmentation types, except on GN and additive SAP noise. The augmented accuracy gain with AT on GN, and SAP noise is mainly due to the fact that in our CA training strategy, GN and SAP noise are seen in every epoch (see Section \ref{subsec:training_protocols}). As expected, the gain in augmented performance on mutually inverse augmentations such as B$+/-$ and S$+/-$ is relatively lower for all the three methods. On the CIFAR-10 dataset, FMA performs better than ST with an average improvement of \textbf{8.94\%}, compared to 8.85\%. On the ImageNet dataset, however, ST performs better than FMA with an average improvement of \textbf{8.27\%} compared to 8.04\%, respectively. On the other hand, AT only improves by an average of \textbf{1.98\%} and \textbf{2.12\%} on the CIFAR-10 and ImageNet dataset, respectively. These results clearly show the effectiveness of our CA training strategy in terms of increasing robustness to multiple augmentations simultaneously while being data-efficient. 

Lastly, we observe the improvement of individual augmentation performance as training progresses for our FMA loss model in Figure \ref{fig:combined_test_individual_CIFAR}. We report highest improvement in the set of \textit{non-inverse} augmentations $\mathcal{A}'$ such as GN, additive SAP noise and GB and relatively lower improvement in the set of \textit{inverse augmentations}. This is primarily because the \textit{non-inverse} augmentations are seen more often due to the training strategy itself when compared to the inverse augmentations such as B$+/-$ and S$+/-$. The performance on clean data is also more or less recovered. 

\section{Conclusion}
\label{sec:Conclusion}
In this work, we proposed a new feature map augmentation (FMA) loss which can be used to efficiently stabilize a CNN to a variety of commonly known input distortions. We also introduced a new combined augmentation (CA) training strategy, which can be used to gain robustness to multiple augmentations \textit{at once} in a data-efficient manner. Using CA, we further improved an existing state-of-the-art method called stability training (ST)~\cite{zheng2016}. In the end, for both CIFAR-10 and ImageNet datasets, we attained a single model each, which have an average augmented accuracy improvement of \textbf{8.94\%} and \textbf{8.04\%} absolute, respectively, while retaining original task performance. In comparison, conventional data augmentation only achieves \textbf{1.98\%} and \textbf{2.12\%}, respectively. These results indicate that \textit{clever combinations of data augmentations, together with additional robustness-focused loss functions, can help improve robustness in a data-efficient manner towards a held-out set of relevant corruptions}. This is \textit{significantly better than conventional data augmentation}.

As a scope of future work, it would be interesting to find underlying similarities between several augmentation types and study their generalization abilities. Based on these clever augmentation sub-sets, FMA loss can be applied on top to attain robust models that also generalize well to unseen augmentations. 

\begin{acks}
The research leading to these results is funded by the German Federal Ministry for Economic Affairs and Energy within the project “KI Absicherung – Safe AI for Automated Driving". The authors would like to thank the consortium for the successful cooperation.
\end{acks}

\bibliographystyle{ACM-Reference-Format}
\bibliography{references}

\end{document}